%% file: _main.tex
\documentclass[runningheads]{llncs}
\usepackage[T1]{fontenc}
\usepackage{graphicx}
\usepackage{hyperref}
\usepackage{color}
\usepackage[ruled,vlined]{algorithm2e}

\usepackage{lineno}
\usepackage{amsmath, amssymb}

\newcommand\btext[1]{{\color{black}{#1}}}
\newcommand{\act}[1]{\textsf{#1}}
\usepackage{booktabs}
\usepackage{multirow}
\usepackage{makecell}

\usepackage[font=small]{caption} 
\usepackage{subcaption} 
\SetAlCapSkip{3pt} 
\usepackage{comment}
\begin{document}
\title{A Rollout-Based Algorithm and Reward Function for Resource Allocation in Business Processes}
\titlerunning{Rollout-based Resource Allocation in Business Processes}
\author{Jeroen Middelhuis\inst{} \and
Zaharah Bukhsh\inst{} \and
Ivo Adan\inst{} \and
Remco Dijkman\inst{}}
\authorrunning{J. Middelhuis et al.}
%
\institute{Eindhoven University of Technology \\
Department of Industrial Engineering and Innovation Sciences
\email{j.middelhuis;z.bukhsh;i.adan;r.m.dijkman@tue.nl}}

\maketitle              

\begin{abstract}
\input{0_abstract}

\end{abstract}
%
%
\section{Introduction} \label{sec:introduction}
\input{1_introduction}

\section{Related work} \label{sec:related_work}
\input{2_related_work}

\section{Background} \label{sec:background}
\input{3_background}

\section{Method} \label{sec:method}
\input{4_method}

\section{Evaluation} \label{sec:results}
\input{5_results}

\section{Conclusion} \label{sec:conclusion}
\input{6_conclusion}

\begin{credits}
\subsubsection{\ackname} The research leading to this paper is supported by the Dutch foundation for scientific research (NWO) under the CERTIF-AI project (nr. 17998).
\end{credits}

%
%
%
\bibliographystyle{splncs04}
\bibliography{references}
\end{document}

%% file: 0_abstract.tex

Resource allocation plays a critical role in minimizing cycle time and improving the efficiency of business processes. 
Recently, Deep Reinforcement Learning (DRL) has emerged as a powerful technique to optimize resource allocation policies in business processes. 
In the DRL framework, an agent learns a policy through interaction with the environment, guided solely by reward signals that indicate the quality of its decisions. However, existing algorithms are not suitable for dynamic environments such as business processes. Furthermore, existing DRL-based methods rely on engineered reward functions that approximate the desired objective, but a misalignment between reward and objective can lead to undesired decisions or suboptimal policies.
To address these issues, we propose a rollout-based DRL algorithm and a reward function to optimize the objective directly. Our algorithm iteratively improves the policy by evaluating execution trajectories following different actions. Our reward function directly decomposes the objective function of minimizing the cycle time, such that trial-and-error reward engineering becomes unnecessary.
\btext{We evaluated our method in six scenarios, for which the optimal policy can be computed, and on a set of increasingly complex, realistically sized process models. The results show that our algorithm can learn the optimal policy for the scenarios and outperform or match the best heuristics on the realistically sized business processes.}

%% file: 1_introduction.tex
Efficient resource allocation plays a crucial role in minimizing the mean cycle time within business processes. In data-driven business process optimization (BPO), methods aim to optimize the process with respect to some objective. Many BPO techniques rely on heuristics or rule-based strategies (e.g., \cite{kuchavr2016automatic,zhao2015optimization}), which are non-adaptive to the dynamicity of business processes. More recently, learning-based methods, such as reinforcement learning (RL), have emerged as a powerful tool for process optimization, leveraging a sequential decision-making framework to allocate resources effectively~\cite{meneghello24,middelhuis_2025}. However, existing RL-based BPO methods rely on engineered reward signals to guide the learning process.

Engineered rewards are designed to guide decision-making agents toward achieving a specific objective. These engineered rewards are approximations of the objective, and maximizing this reward should also optimize the objective. However, engineered rewards, when misspecified, can lead to unintended agent behavior~\cite{amodei2016concrete}. One risk is reward hacking, where an agent exploits loopholes in the reward function, resulting in unintended or undesirable behaviors without completing the intended objective~\cite{pan2022the}. For example, a cleaning robot may be given points for picking up trash. Instead of cleaning up trash, it can also knock over trash, thus maximizing the reward it earns, but failing to achieve the objective of cleaning an area. Similar unintended behaviors can arise in business processes where an agent optimizes a different objective than intended.

In the domain of BPO, three different reward functions for cycle time minimization have been proposed, but each has inherent limitations that can lead to suboptimal policies. The first reward function rewards an agent +1 for each case completion but does not account for the realized cycle time of the case~\cite{lo2023action,zbikowski_deep_2023}. The second reward function penalizes the agent equivalent to the cycle time of a completed case~\cite{huang2011reinforcement}. However, this approach may inadvertently incentivize the agent to delay or avoid completing cases to minimize penalties. The third reward function returns the inverse of the cycle time when a case completes, ensuring that fast case completions yield a higher positive reward~\cite{meneghello24,middelhuis_2025}. However, this reward function can be gamed by the agent, resulting in suboptimal policies. While these methods can lead to efficient policies for certain processes, they also require problem-specific engineering of reward functions and algorithms, leading to suboptimal policies or failure to learn a policy. Furthermore, the typical DRL algorithms, such as proximal policy optimization (PPO)~\cite{schulman2017proximal}, used in the aforementioned studies are not specifically designed to handle dynamic environments, such as a business process, making it challenging to learn effective policies~\cite{TEMIZOZ2025}.

This paper introduces a rollout-based DRL algorithm and a reward function to optimize the mean cycle time. Our algorithm directly determines the sum of rewards following different actions, removing the need to approximate state-action values. Our reward function eliminates the need for extensive reward engineering, ensuring that the learned policies directly optimize the desired objective. We provide a comprehensive evaluation of our algorithm and benchmark it with existing reward functions using the state-of-the-art PPO algorithm~\cite{schulman2017proximal}. The results show that our algorithm successfully learns the optimal policy across for typical business process pattenrs and outperforms the PPO algorithm.

The remainder of the paper is structured as follows. Section~\ref{sec:related_work} provides an overview of the related work. Section~\ref{sec:background} introduces background concepts. Section~\ref{sec:method} explains our algorithm and reward function. Section~\ref{sec:results} evaluates our method and Section~\ref{sec:conclusion} concludes our work.




%% file: 2_related_work.tex
This section reviews existing algorithms and reward functions for resource allocation in business processes, highlighting their limitations and motivating the gap in the literature. Section~\ref{sec:rel_algorithms} discusses existing algorithms and identifies their shortcomings for business processes. Section~\ref{sec:engineered_rewards} details existing reward functions and highlights the need for a more effective reward function.

\subsection{Algorithms for business process optimization} \label{sec:rel_algorithms}
BPO algorithms are designed to improve overall process efficiency by optimizing key performance indicators~\cite{kubrak2022prescriptive}. Several BPO methods rely on static resource rankings based on event logs~\cite{arias2016framework,kuchavr2016automatic,zhao2015optimization} or process models~\cite{schumann2024optimizing,si2018petri}. While these methods offer a structured approach, they lack adaptability in dynamic environments.

Dynamic approaches assign resources at runtime using process state data. Park and Song~\cite{park2019prediction} employ predictive modeling to minimize costs. Recently, RL has emerged as a powerful tool for BPO due to its ability to adapt to observed process dynamics. Early work applied Q-learning to minimize cycle time~\cite{huang2011reinforcement}. However, Q-learning struggles with scalability as it stores each state-action pair separately, making it inefficient for large-scale, continuous business environments. DRL deals with this issue with approximators~\cite{sutton2018reinforcement} and has been used for optimizing business processes~\cite{lo2023action,lo2024universal,zbikowski_deep_2023}, demonstrating effectiveness in large-scale processes~\cite{meneghello24,middelhuis_2025,neubauer2022resource}.

Popular DRL algorithms have been successfully applied in game environments~\cite{mnih2013playing,schulman2017proximal}. However, these environments differ from BPO settings, as they often have well-defined, deterministic rules. For instance, in chess, a move has a predictable and deterministic outcome. In contrast, business processes are driven by random events, such as unpredictable case arrivals and variable activity durations. Standard DRL algorithms are not explicitly designed to handle such stochastic environments~\cite{TEMIZOZ2025}. Within the BPO domain, applications of these algorithms often do not converge to optimal policies~\cite{huang2011reinforcement,meneghello24,middelhuis_2025,zbikowski_deep_2023}. To address this issue, we propose a rollout-based algorithm to learn optimal policies for business processes. \btext{Our approach is inspired by other rollout-based algorithms~\cite{lagoudakis2003reinforcement,TEMIZOZ2025} and we extended this idea to fit the requirements of resource allocation for business processes. For a more extensive review of the resource allocation algorithms, we refer to Pufahl et al.~\cite{pufahl2025resource}.}

\subsection{Reward functions for business process optimization}\label{sec:engineered_rewards}
While DRL offers a promising framework for resource allocation, its success depends on the design of the reward function, which remains a key challenge in BPO. An effective reward function should provide a feedback signal that guides the agent toward optimizing a specific objective function.

In BPO, three reward functions have been proposed for cycle time minimization: (1) a reward of +1 per completed case~\cite{lo2023action,zbikowski_deep_2023}; (2) a penalty equal to cycle time per completed case~\cite{huang2011reinforcement,neubauer2022resource}; and (3) a reward of $\frac{1}{1+CT}$ per completed case, where $CT$ is cycle time~\cite{meneghello24,middelhuis_2025}. While these reward functions encourage cycle time minimization, they are prone to reward hacking~\cite{amodei2016concrete,pan2022the}, in which the agent optimizes a different objective than intended.

The first reward function encourages case completions but ignores the temporal aspect, which may lead to inefficiency. The second reward function more directly targets the cycle time but only returns penalties. Consequently, avoiding case completions by, for example, postponing, results in zero rewards, which are better than penalties. The third reward function rewards short cases disproportionately, which can lead to unintended behavior. For example, completing two cases with cycle times $\{1,10\}$ yields a higher cumulative reward than two cases with cycle times $\{5,5\}$ (a cumulative reward of $\frac{13}{22}$ compared to $\frac{1}{3}$, respectively), despite the latter having a shorter mean cycle time. While a postpone penalty was introduced~\cite{middelhuis_2025} to mitigate this issue, it required additional engineering and did not fully eliminate reward hacking. 

A broader limitation shared by all three reward functions is the sparse and delayed nature of the reward function. The agent only receives a reward after a case is completed, and intermediate actions receive no rewards, making learning effective policies challenging~\cite{arjona2019rudder}. These limitations of the existing engineered rewards highlight the need for a reward function that guides the agent toward cycle time minimization and does not require problem-specific reward engineering.

In this paper, we introduce a reward function that decomposes the contribution to the sum of cycle times and integrates it with a rollout-based DRL algorithm to learn optimal resource allocation policies.

%% file: 3_background.tex
This section introduces the resource allocation problem and how it can be formulated and solved as an MDP. Section~\ref{sec:bpo_concepts} defines resource allocation concepts. Section~\ref{sec:mdp_drl_concepts} presents the Markov Decision Process (MDP) framework and how DRL can be applied to learn a resource allocation policy on an MDP.

\subsection{Resource allocation in business processes} \label{sec:bpo_concepts}
\begin{figure}
    \centering
    \includegraphics[width=0.5\linewidth]{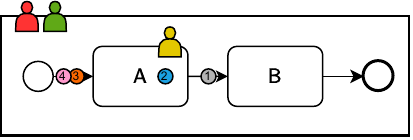}
    \caption{Example of a process model with two activities, three resources, and four active cases. Activities \act{A} and \act{B} have unassigned activity instances ($K_\act{A} = \{3,4\}, K_\act{B} = \{1\}$). The \texttt{yellow} resource is processing activity instance 2 ($\mathcal{R}^-=\{r_{\texttt{yellow}}\}$, $B=\{(2,r_{\texttt{yellow}})\}$). The \texttt{red} and \texttt{green} resources are available ($\mathcal{R}^+ = \{r_{\texttt{red}}, r_{\texttt{green}}\}$).}
    \label{fig:process_model}
\end{figure}

In a business process, cases from a set $\mathcal{C}$ arrive according to some arrival process. A case consists of a series of activities from a set $\mathcal{A}$ that must be executed by resources from a set $\mathcal{R}$. When a specific case $c \in \mathcal{C}$ enters the process, it is added to the set of ongoing cases $C \subseteq \mathcal{C}$ and initiates an activity instance $k \in \mathcal{K}$. An activity instance represents an execution of an activity for a specific case, and a case may involve multiple such instances throughout its lifecycle. Subsequent activity instances are generated during the execution of the case. In this paper, we assume a Poisson arrival process and exponential processing times. Fig.~\ref{fig:process_model} shows an example of a business process.

Activity instances that are awaiting execution form the set of unassigned activity instances, denoted as $K \subseteq \mathcal{K}$. For a specific activity, $act \in \mathcal{A}$, the subset of activity instances and unassigned activity instances are denoted by $\mathcal{K}_{act} \subseteq \mathcal{K}$ and $K_{act} \subseteq \mathcal{K}_{act}$, respectively. To execute an unassigned activity instance $k \in K$, a resource $r \in \mathcal{R}$ must be allocated. The resource set $\mathcal{R}$ is divided into available resources $\mathcal{R}^+ \subseteq \mathcal{R}$ and unavailable resources $\mathcal{R}^- \subseteq \mathcal{R}$, where $\mathcal{R}^+ \cup \mathcal{R}^- = \mathcal{R}$ and $\mathcal{R}^+ \cap \mathcal{R}^- = \emptyset$. Not all resources are qualified to perform every activity, often due to varying expertise or authorization levels. Therefore, for each activity, $act \in \mathcal{A}$, the set of eligible resources, as defined by the process model, is $\mathcal{R}_{act} \subseteq \mathcal{R}$.

Using these components, we can form a resource-to-activity assignment $(r, act)$. Considering resource eligibility, we define the set of allowed assignments as $\mathcal{D}=\{(r, act) \mid act \in \mathcal{A}, r \in \mathcal{R}_{act}\}$. We define the set of possible assignments at runtime as $D=\{(r, act) \mid (r,act) \in \mathcal{D}, r \in \mathcal{R}^+, |K_{act}|>0\}$. Making an assignment $(r, act)$ starts the execution of $(r, k)$, which is the assignment of resource $r \in R_{act}$ to an activity instance $k \in K_{act}$. \btext{Then, $r$ and $k$ are taken from $\mathcal{R}^+$ and the set of assigned activity instances $K$, and instead put into $\mathcal{R}^-$ and set of assigned activity instances $Z$, and the set of ongoing assignments $B \subseteq Z \times \mathcal{R}^-$.}

The current condition of the process can be represented as an execution state $(C, K, \mathcal{R}^+, \mathcal{R}^-, B, t)$, where $t \in T \subseteq [0, \infty)$ is a moment in time. $\mathcal{S}$ is the set of all possible execution states. In each state, the agent can decide either to make an assignment $(r, act) \in \mathcal{D}$ or to postpone the decision. After making a decision, the state transitions to the next state $(C^\prime, K^\prime, \mathcal{R}^{+\prime}, \mathcal{R}^{-\prime}, B^\prime, t^\prime)$ in two steps, as follows. In the first step, if the agent makes an assignment $(r, act)$, a resource $r$ is assigned to an unassigned activity instance $k \in K_{act}$, i.e., $(r,k)$. Next, changes are made to the resource sets $\mathcal{R}^{+\prime}=\mathcal{R}^{+} - \{r\}$ and $\mathcal{R}^{-\prime}=\mathcal{R}^{-} \cup \{r\}$, the set of unassigned activity instances $K^\prime=K-\{k\}$, and the set of ongoing assignments $B'=B\cup\{(r, k)\}$. If the agent chooses to postpone, nothing happens in this step. In the second step, an event is triggered. This can either be the arrival of a new case or the completion of an assigned activity instance $ k\in Z$. We denote the set of events that is possible in a state \btext{$s$} after executing action $a \in D \cup \{\mathrm{postpone}\}$ as $E(s,a)$. These events all have a timing probability distribution associated with them, which models how long it will take before they occur (e.g., how long it will take from an action until the next case arrival). Since we assume exponential distributions, the timing distribution of event $e\in E(s,a)$ can be represented by a rate $\lambda_e$. When a new case  $c^\prime$ arrives, this also generates an initial activity instance $k^\prime$, such that \btext{$C^\prime = C \cup \{c^\prime\}$} and \btext{$K^\prime = K \cup \{k^\prime\}$}. When an assignment $(r, k)$ completes, the state changes such that $B'=B-\{(r, k)\}$, $\mathcal{R}^{+\prime}=\mathcal{R}^{+} \cup \{r\}$ and $\mathcal{R}^{-\prime}=\mathcal{R}^{-} - \{r\}$. Subsequently, a new activity instance $k^\prime$ may be generated depending on the process model, which changes $K^\prime = K \cup k^\prime$. If the completed activity instance results in the completion of a case, this case is removed from the set of active cases $C^\prime = C - \{c\}$. The cycle time of a case $CT_c$ is equal to the difference between its arrival and departure time.

\subsection{Deep Reinforcement Learning} \label{sec:mdp_drl_concepts}
We aim to solve the resource allocation problem using Deep Reinforcement Learning (DRL), which combines deep neural networks with reinforcement learning (RL) to solve complex decision-making tasks. In RL, an agent interacts with an environment, taking actions based on a policy $\pi(a|s)$ to maximize cumulative rewards~\cite{sutton2018reinforcement}. The agent observes a state $s$, selects an action $a$, receives a reward $r$, and transitions to a new state $s^\prime$. We refer to this interaction as a decision step. DRL leverages deep learning to approximate policies, value functions $V^\pi(s)$, or Q-values, such that it can efficiently handle large state spaces. To apply DRL, we must model resource allocation as a Markov Decision Process (MDP).

An MDP provides a mathematical framework for sequential decision-making~\cite{sutton2018reinforcement}, where outcomes are partly under the control of an agent and partly caused by randomness. Formally, an MDP is a tuple $(S, A, P, R)$, where $S$ is a finite set of states, $A$ a finite set of actions, $P(s^\prime|s,a)$ is the probability that, when an agent takes an action $a$ in state $s$, the environment transitions into the state $s^\prime$, and $R(s,a)$ is function that models the reward that the agent gets when performing action $a$ in state $s$. Guided by the reward, The agent learns a policy $\pi(a|s)$ to maximize cumulative rewards over an episode, which is a sequence of interactions between the agent and the environment, during which the agent decides on an action $a$, and the environment subsequently applies the action to transition into a next state $P(s^\prime|s,a)$ and gives a reward $R$. The episode starts in an initial state $s_0$ and ends in a terminal state.

Standard MDPs assume decisions occur at discrete intervals, whereas decision-making in business processes often occurs at random intervals. To address this, a Continuous-Time MDP (CTMDP) extends the MDP framework by allowing decision steps to occur at arbitrary time points.


%% file: 4_method.tex
This section presents our proposed algorithm to learn optimal resource allocation policies using DRL. Since DRL uses an MDP model to learn, we first explain in Section~\ref{sec:bp_as_MDP} how a resource allocation problem in a business process, as it is described in Section~\ref{sec:bpo_concepts}, can be mapped to a (CT)MDP. Subsequently, Section~\ref{sec:proposed_algorithm} introduces our proposed rollout-based algorithm to learn the optimal policy. The learning process is guided by our reward function, which is equivalent to the objective, which is described in Section~\ref{sec:proposed_reward_function}. Finally, we need to compute the optimal policy to ensure that our learned policies are optimal. Section~\ref{sec:optimal_policy} describes how we can transform the CTMDP, which is a business process, into a discrete-time MDP such that we can compute the optimal policy.

\subsection{The business process as a CTMDP} \label{sec:bp_as_MDP}
\begin{figure}
    \centering
    \includegraphics[width=0.7\linewidth]{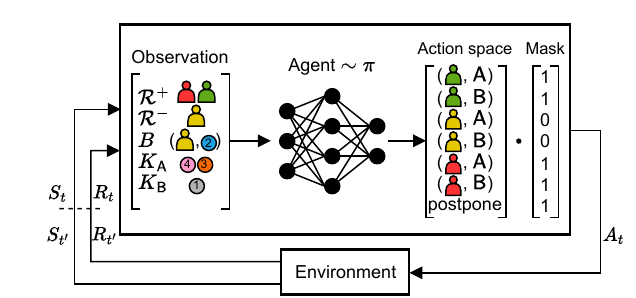}
    \caption{The agent-environment interaction for the resource allocation problem based on the business process in Fig.~\ref{fig:process_model}.}
    \label{fig:rl_loop}
\end{figure}

To apply a DRL algorithm to solve the resource allocation problem described in Section~\ref{sec:bpo_concepts}, we map it to a CTMDP $(S, A, P, R)$. An overview of the RL based on the (CT)MDP is shown in Fig.~\ref{fig:rl_loop}. In the remainder of this section, we detail how each component $S, A, P$ of the CTMDP is created from the process. Since we propose both an algorithm and reward function, we provide an in-depth overview of the reward function $R$ in Section~\ref{sec:proposed_reward_function}.

\paragraph{\normalfont \textbf{State.}}
The state contains all information related to the condition of the business process, such as the waiting activity instances and assignments that are being processed. Providing the agent with every possible state as an input would cause a state explosion, making the problem computationally infeasible. For that reason, at each decision step, the agent partially observes the state based on features that are relevant to the decision and are derived from the state variables. The information represented by these features is called an observation, which we model as follows:

\vspace{-0.7em}

\begin{itemize}
    \item A binary mapping $\mathcal{R} \rightarrow \{0,1\}$, indicating whether the resource is available or not. For resource $r \in \mathcal{R}$, this feature equals 1 if $r \in \mathcal{R}^+$ and 0 if $r \in \mathcal{R}^-$.
    \item A binary mapping $\mathcal{D} \rightarrow \{0,1\}$, indicating whether an assignment $(r,act) \in \mathcal{D}$ is being executed or not. For assignment $(r, act) \in \mathcal D$, this feature equals 1 if  $\exists r \in \mathcal{R}^+,k \in K_{act}$ such that $(r, k) \in B$, and 0 otherwise.
    \item A continuous mapping $\mathcal{A} \rightarrow [0,1]$, indicating the number of waiting cases at activity $act \in \mathcal{A}$. Similarly to \cite{middelhuis_2025}, we scale this feature between 0 and 1 by dividing it by 100 and truncating values above 1. For activity $act \in \mathcal{A}$, this feature equals $\min(\frac{|K_{act}|}{100}, 1)$.
\end{itemize}


\paragraph{\normalfont \textbf{Action.}}
Based on the observation of the state, the agent selects an action from the action space, which includes all possible resource-to-activity assignments $\mathcal{D}$, as well as a postpone action. Postponing to wait for better future decisions can be beneficial when only inefficient assignments are currently available. In some states, no assignments are possible, making postponing the only option. During training, we differentiate between forced and voluntary postponing, allowing the agent to learn when postponing is a strategic choice rather than a necessity.

At each decision step, we check which actions are possible based on the state, and infeasible actions are masked, disallowing the agent to choose these actions. For example, assignments that do not have their respective resource or unassigned activity instances available are masked.


\paragraph{\normalfont \textbf{State transition.}}
As explained in Section~\ref{sec:background}, when taking an action $a\in D \cup \{\mathrm{postpone}\}$ in state $s\in S$, the process transitions into a state $s'\in S$, depending on the event $e\in E(s,a)$ from the set of possible events, which happens after executing the action with a rate $\lambda_e$. To represent this in the CTMDP, we define $P(s'|s, a)$, which specifies the probability of transitioning from $s$ to $s'$ after taking action $a$. In the CTMDP, the time until the next decision step is exponential with rate $\sum_{e\in E(s,a)} \lambda_e$ and is independent of the event that triggers it. The probability that it reaches state $s'$ due to event $e'\in E(s,a)$ is:

\vspace{-0.5em}

\begin{equation}
    \frac{\lambda_{e'}}{\sum_{e\in E(s,a)} \lambda_e}
\end{equation}

The CTMDP formulation outlined in this section enables us to train a DRL agent to optimize resource allocation policies.

\subsection{Rollout-based algorithm} \label{sec:proposed_algorithm}
The algorithm that we use to find the best policy for assigning resources to tasks is based on the rollouts of that policy. A rollout refers to the process of simulating an agent's interaction with the environment (i.e., repeatedly performing an action and checking its effect) while following a specific policy for a specific number of decision steps. The benefit of using rollouts is that we can use the sum of rewards, which is also called the return, to represent exactly the objective we aim to optimize, such as the sum of cycle times. We use a rollout to estimate the expected return of a policy, given by: 

\vspace{-0.5em}

\begin{equation}
    G_t = \sum_{i=0}^{N-1}R_{t+i+1}
\end{equation}

\noindent where $R_j = R(s_j, a_j)$ is the reward of decision step $j$, and $N$ represents the number of decision steps in each rollout. To improve the reliability of the estimate, we perform $M$ rollouts and compute the average return. We use common random numbers to reduce variance across rollouts actions~\cite{TEMIZOZ2025}. Algorithm~\ref{alg:rollouts} presents our rollout-based approach.

\begin{algorithm}[h]
\caption{Rollout-based algorithm}\label{alg:rollouts}

\KwIn{$\mathcal{M}, \pi_0, I, M, N$}
\KwOut{$\pi$}

\nl $\pi = \pi_0$\;
\nl $V^\pi(s_0) \gets \text{\textbf{evaluate}}(\mathcal{M}, \pi)$\;

\nl \For{$i=1,2,\dots,I$}{
    \nl Generate states $\mathcal{S}$ from $\mathcal{M}$\;
    \nl dataset = $\emptyset$\;
    \nl \ForEach{$s \in \mathcal{S}$}{
        \nl \ForEach{$a \in A(s)$}{
            \nl $G^\pi(s,a) \gets \text{\textbf{rollout}}(\mathcal{M}, \pi, M, N)$\;
        }
        \nl $a^*=\displaystyle\arg\max_{a \in A(s)} G^\pi(s,a)$\;
        \nl dataset $\gets$ dataset $\cup \{(s, a^*)\}$\;
    }
    \nl $\pi^\prime \gets \text{\textbf{learn}}(\pi, \text{dataset})$\;    
    \nl $V^{\pi^\prime}(s_0) \gets \text{\textbf{evaluate}}(\mathcal{M}, \pi^\prime)$\;
    \nl \If{$V^{\pi^\prime}(s_0) > V^\pi(s_0)$}{
        \nl $\pi = \pi^\prime$\;
    }
}
\nl Return $\pi$\;
\end{algorithm}

Our rollout-based algorithm has three steps: \emph{initialization}, \emph{evaluation}, and \emph{improvement}. In the \emph{initialization} step \btext{(line 1)}, we define a bootstrap policy $\pi_0$, for which we use a greedy policy. During the \emph{evaluation} step \btext{(lines 2, 12-14)}, we deploy $\pi$ on the business process and evaluate its performance. In the \emph{improvement} step \btext{(lines 3-11)}, we explore alternative actions and update the policy to $\pi^\prime$. The evaluation and improvement steps are repeated for $I$ iterations.

\begin{figure}
    \centering
    \includegraphics[width=0.6\linewidth]{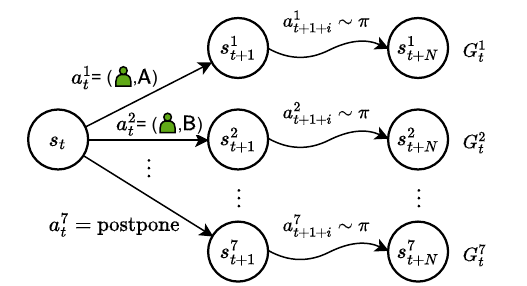}
    \caption{Rollouts for the example in Fig.~\ref{fig:rl_loop}\btext{, representing lines 7-10 of Algorithm~\ref{alg:rollouts}}. For each action $a \in A(s_t)$, a rollout is performed. Consequently, the state transitions to state $s_{t+1}$. Next, a rollout of $N-1$ decision steps following policy $\pi$ is performed to estimate the return $G_t$ of each action.}
    \label{fig:rollouts}
\end{figure}

During the improvement step, we use rollouts to estimate the best action to take in a state, i.e., the action that has the highest average return. First, we sample a set of random states $\mathcal{S} \subseteq S$. For each state $s \in \mathcal{S}$, we determine the optimal action $a^*$  from the set of possible actions $a \in A(s)$ in that state using rollouts. The (CT)MDP $\mathcal{M}$ provides the framework for generating the random states and performing the rollouts. Figure~\ref{fig:rollouts} illustrates how rollouts assess different actions.

The best state-action pairs identified through rollouts are used to update the policy. To represent the policy, we use a neural network where the input represents the observed state, and the output corresponds to the best action label. After updating the policy $\pi^\prime$, we evaluate $V^{\pi^\prime}(s_0)$, where $s_0$ in the initial state of $\mathcal{M}$, which is an empty process. If the new policy performs better, we proceed with $\pi = \pi^\prime$. Otherwise, we revert to the previous policy $\pi$.

\subsection{Reward function} \label{sec:proposed_reward_function}

\begin{figure}
    \centering
    \includegraphics[width=0.7\linewidth]{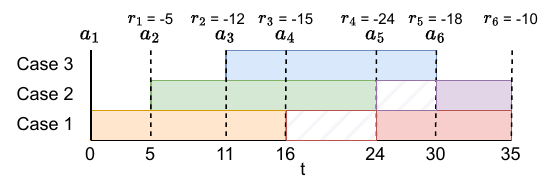}
    \caption{\btext{Example of the proposed reward function for three cases. At each decision step, the reward function returns a reward equal to the contribution to the total cycle time during the state transition. Waiting cases (hatched area) also incur cycle time. The reward is returned after the state transition.}}
    \label{fig:reward_example}
\end{figure}

A fundamental component of the DRL framework is the reward function, which serves as the guiding signal for the agent to learn an optimal policy. If the reward signal does not accurately reflect the objective function, the agent may learn suboptimal policies. We propose a reward function that, when maximized, directly minimizes the total cycle time without requiring reward engineering. The reward function is dense as it provides a reward at each decision step, which reduces the variance in rewards and improves learning efficiency~\cite{arjona2019rudder}. The reward for an action taken in state $s$, which occurs at time $t$ is defined as:

\vspace{-0.5em}

\begin{equation}
    r_t = -|C|(t^\prime - t)
\end{equation}

\noindent where $|C|$ is the number of active cases in the process at $t$. In this equation, $t$ and $t^\prime$ and the times of the current and next decision step, respectively (i.e., according to $s$ and $s^\prime$). We give a negative reward such that the sum of cycle times is minimized when the cumulative reward is maximized. Fig.~\ref{fig:reward_example} shows an example of how the reward function is used for three cases consisting of one or two activities.

In this example, at $t=0$, case 1 arrives and is assigned by taking action $a_1$. The state transitions due to the arrival event of case 2. Since five time units passed and there there is one case in the process, a reward of $r_1=-5$ is returned. Similarly, for instance, for $a_4$, eight time units pass before transitioning to the next state while there are three cases in the system, resulting in a reward of $r_4=-24$. Using this reward function, the agent receives a continuous feedback signal equal to the contribution to the sum of cycle times during that transition. 

Since the state transition time cannot be minimized, optimizing our reward function effectively reduces $|C|$, the number of cases in the system. Our reward function explicitly penalizes high values of $|C|$, and with a constant arrival rate $\lambda$, minimizing $|C|$ directly leads to a lower cycle time, as $\hat{CT} = \hat{C}/\lambda$, according to Little's law~\cite{hopp2011factory}, where $\hat{C}$ is the average number of cases in the system. Thus, by minimizing $|C|$, we naturally minimize the mean cycle time $\hat{CT}$.

\subsection{Uniformization of the CTMDP and optimal policy} \label{sec:optimal_policy}
To assess if our algorithm can learn the optimal policy, we first have to compute this policy. This can feasibly be done for small models - like the ones we will use in the evaluation - as described in this section.

Computing the optimal policy of a CTMDP using value iteration is not straightforward due to the stochastic transition times on which the rewards depend. Therefore, we first introduce a transformation to uniformize the CTMDP into a discrete-time MDP~\cite{tijms1986stochastic}. We can then apply the value iteration algorithm to the MDP to compute the optimal policy. This policy is also optimal for the CTMDP in terms of average reward~\cite{tijms1986stochastic}. \btext{Note that the MDP is only required to compute the optimal policy and is not required for our algorithm presented in Section~\ref{sec:proposed_algorithm}.} First, we choose $\tau$ to represent the time between states, such that $t \in T = \{n\tau \mid n \in \mathbb{N}_0\}$:

\vspace{-0.5em}

\begin{equation}
    0 < \tau \leq \min_{s,a} \tau_s(a) \text{  where  } \min_{s,a} \tau_s(a)= \frac{1}{\max_{s,a}\sum_{e\in E(s,a)} \lambda_e} \label{eq:tau}
\end{equation} 

\noindent where $\min_{s,a} \tau_s(a)$ is the minimum expected transition time in the CTMDP. Based on $\tau$, we can transform the CTMDP into a discrete-time MDP with the same state and action space and the following one-step rewards and one-step transition probabilities: 

\vspace{-1em}

\begin{align}
    & \tilde{r}_s(a) =  \frac{r_s(a)}{\tau_s(a)} && s \in S \textrm{ and } a \in A(s) \label{eq:average_cost}\\
    & \tilde{p}_{ss^\prime}(a)=
        \begin{cases}
            \frac{\tau}{\tau_s(a)}p_{ss^\prime}(a), & s \neq s^\prime,\\
            \frac{\tau}{\tau_s(a)}p_{ss^\prime}(a) + (1 - \frac{\tau}{\tau_s(a)}), & s = s^\prime,
        \end{cases} && s, s^\prime \in S \textrm{ and } a \in A(s)\label{eq:transition_probabilities}
\end{align}

\noindent where $r_s(a)$ and $\tau_s(a)$ are the expected reward and expected time in state $s$ when taking action $a$ in the CTMDP, respectively. Furthermore, $p_{ss^\prime}(a)$ is the expected transition probability of going from state $s \in S$ to state $s^\prime \in S$ in the CTMDP. Due to the non-uniformity of the transition time in the CTMDP, expected rewards are transformed into reward rates, and transition probabilities include fake transitions to make the times in between decision steps uniform.

Using value iteration, we compute the optimal value function $V^*$ for the MDP. However, due to unbounded queue length features, the state space is infinite. Therefore, we introduce an upper bound of 100 on the queue length features, with substantial penalties for transitions that access states beyond this limit, guiding the agent toward more favorable states. We follow the value iteration algorithm from Tijms~\cite{tijms1986stochastic}.. Since we do not discount future rewards, the value function converges when the difference between the minimum and maximum absolute delta between two successive value functions converges to zero.

By applying value iteration, we obtain the \textit{optimal policy} for the bounded MDP. While this formulation introduces an approximation, the upper bound is set sufficiently high to ensure it is never exceeded under a random policy in our evaluations (Section~\ref{sec:evaluation_protocol}). As a result, the optimal policy derived from the bounded MDP is unlikely to differ from that of the unbounded MDP. Therefore, we refer to the policy obtained via value iteration on the bounded MDP as the \textit{optimal policy}.

%% file: 5_results.tex
This Section evaluates our approach. Section~\ref{sec:evaluation_protocol} introduces our evaluation protocol. Section~\ref{sec:evaluation} demonstrates the ability of our method to learn optimal policies. Section~\ref{sec:evaluation_benchmark_rewards} shows the performance of different reward functions and Section~\ref{sec:composite_evaluation} shows an evaluation on more realistically sized models.

\subsection{Evaluation setup}\label{sec:evaluation_protocol}
\begin{figure}
     \centering
     \begin{subfigure}[b]{0.45\textwidth}
         \centering
         \includegraphics[width=\textwidth]{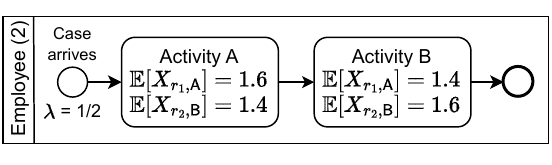}
         \caption{Scenario 1: Low utilization}
         \label{fig:low_utilization}
     \end{subfigure}
     \hspace{0.02\textwidth}
     \begin{subfigure}[b]{0.45\textwidth}
         \centering
         \includegraphics[width=\textwidth]{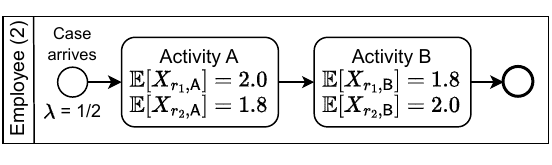}
         \caption{Scenario 2: High utilization}
         \label{fig:high_utilization}
     \end{subfigure}
     
     \begin{subfigure}[b]{0.45\textwidth}
         \centering
         \includegraphics[width=\textwidth]{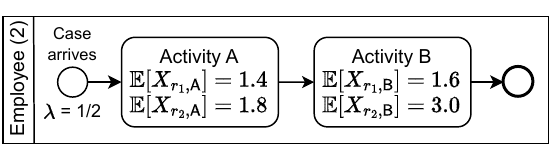}
         \caption{Scenario 3: Slow server}
         \label{fig:slow_server}
     \end{subfigure}
     \hspace{0.02\textwidth}    
     \begin{subfigure}[b]{0.45\textwidth}
         \centering
         \includegraphics[width=\textwidth]{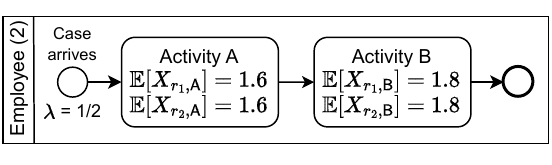}
         \caption{Scenario 4: Slow downstream}
         \label{fig:slow_downstream}
     \end{subfigure}

     \begin{subfigure}[b]{0.45\textwidth}
         \centering
         \includegraphics[width=\textwidth]{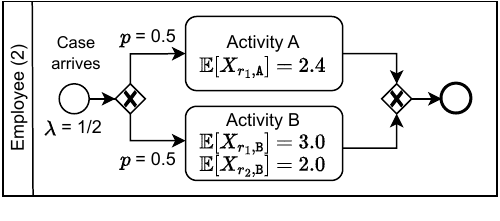}
         \caption{Scenario 5: N-network}
         \label{fig:n_network}
     \end{subfigure}
     \hspace{0.02\textwidth}
     \begin{subfigure}[b]{0.45\textwidth}
         \centering
         \includegraphics[width=\textwidth]{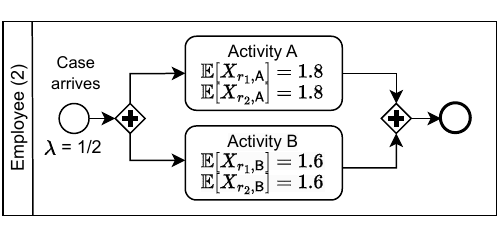}
         \caption{Scenario 6: Parallel}
         \label{fig:parallel}
     \end{subfigure}

     \caption{Six scenarios with two activities and resources. In each scenario, cases arrive with a rate of $\lambda=1/2$. Activity $act$ is executed by resource $r$ with rate $1/\mathbb{E}[X_{r,act}]$.}
     \label{fig:scenarios}
\end{figure}

For the evaluation, we use the scenarios from Middelhuis et al.~\cite{middelhuis_2025} shown in Figure~\ref{fig:scenarios}, which represent typical process patterns.

Similar to Middelhuis et al.~\cite{middelhuis_2025}, we simulate 2500 cases per episode, which terminates when all cases are completed. For evaluation we use 300 episodes. Before training and evaluating our method, we tune hyperparameters on the MDP through grid search. The selected parameter values are applied when training all policies. We tune the number of rollouts per action, $M \in \{50, 100, 200\}$, and rollout length, $N \in \{25, 50, 100\}$, and the transformation parameter $\kappa \in \{0.1, 0.25, 0.5, 1\}$, which determines the MDP step size $\tau = \kappa \min_{s,a} \tau_s(a)$. To ensure consistency between MDP and CTMDP settings, we scale the rollout length $N$ with $\tau$. Hyperparameter tuning leads to a value of $M = 100$. A longer rollout length of $N = 100$, and a $\kappa = 0.5$, aligning with the recommendation of Tijms~\cite{tijms1986stochastic}. For the neural network architecture, we use two layers of 128 neurons. For each model, we generate 5000 random states in each iteration and perform 10 training iterations, of which we save the best model.


\subsection{Evaluation results of the proposed method}\label{sec:evaluation}
We train on both the CTMDP and MDP versions of the six scenarios and evaluate each method on the CTMDP, as it represents the original business process. \btext{While the transformation to MDP is not required to use our proposed method, we train an agent on the MDP to investigate if the environment with more predictable rewards impacts the learning process.} 

Table~\ref{tab:results_our_method} shows the results, comparing our method, trained on both CTMDP and MDP, to the optimal policy and three heuristics: Shortest Processing Time (SPT), First-In-First-Out (FIFO), and a random policy. \btext{We use a t-test ($\alpha$ = 0.05) to assess statistical significance. Results not significantly different from the optimal policy shown in bold. For each method, we show the optimality gap: the percentage difference to the optimal policy.}

\begin{table}
\centering
\caption{Mean cycle time of the optimal policy and optimality gap (\%) between the optimal policy and other methods, including our approach and benchmark heuristics.}
    \begin{tabular}{lcccccc}
    \toprule
    \multirow{2}{*}{Scenario}         & \multirow{2}{*}{\makecell{Optimal \\ policy}}               & \multicolumn{2}{c}{\btext{Our method}} & \multirow{2}{*}{SPT}   & \multirow{2}{*}{FIFO}  & \multirow{2}{*}{Random}  \\ \cmidrule{3-4}
                     &                & CTMDP      & MDP       &       &       &        \\ \midrule
    Low Utilization  & 5.7 (0.05)   & \textbf{-0.41\%} & \textbf{1.03\%} & 4.05\%  & 5.05\%  & 17.78\%  \\
    High Utilization & 18.0 (0.55)  & \textbf{1.31\%} & \textbf{2.40\%} & 9.10\%  & 49.23\%  & 100.52\% \\
    Slow Server      & 11.7 (0.23)  & \textbf{-0.47\%} & \textbf{1.60\%} & 151.14\% & 58.57\%  & 94.90\%  \\
    Slow Downstream  & 10.1 (0.20)  & \textbf{-1.29\%} & \textbf{-0.34\%} & 46.79\%  & \textbf{-0.51\%} & 14.94\%  \\
    N-system         & 5.8 (0.06)   & \textbf{-1.33\%} & \textbf{-1.15\%} & 23.88\%  & \textbf{1.31\%}  & 11.71\%  \\
    Parallel         & 9.7 (0.19)   & \textbf{1.56\%} & \textbf{1.06\%} & 44.00\%  & \textbf{-2.68\%} & 16.41\%  \\ 
    \bottomrule
    \end{tabular}
\label{tab:results_our_method}
\end{table}

Table~\ref{tab:results_our_method} demonstrates that our method successfully learns the optimal policy across all scenarios when trained on either CTMDP or MDP. \btext{We trained on both the CTMDP and MDP to investigate if the variable time between decision steps affects performance. However, no significant differences were found.} Interestingly, the solution for the FIFO policy in the parallel scenario seems to outperform the optimal policy, though not significantly. This difference occurs because the optimal solution is derived based on the features in our observations, whereas the FIFO policy directly allocates resources according to the state.

In the parallelism scenario, the optimal policy involves executing an activity of a case that already has one of its two activities completed, thereby completing the case with the next activity. Due to the nature of our observation, the agent is unable to learn this specific strategy. Consequently, the optimal policy is optimal for our representation of the (CT)MDP, which is learned by our method.

\subsection{Evaluation results of PPO with different reward functions}\label{sec:evaluation_benchmark_rewards}
PPO is widely used to train DRL agents for resource allocation in business processes~\cite{lo2023action,lo2024universal,meneghello24,middelhuis_2025}. However, existing PPO-based methods are trained under varying settings. In this section, we compare PPO’s performance using different existing reward functions and our proposed reward function.

Each existing reward function gives a reward when case $c \in \mathcal{C}$ is completed: (a) +1, (b) $-CT_c$, and (c) $\frac{1}{1+CT_c}$. In state transition following actions that did not result in a case completion, a reward of zero is returned. A PPO agent combined with reward function (c) is considered the current state-of-the-art method for resource allocation in business processes~\cite{meneghello24,middelhuis_2025}.

We use the hyperparameters from Middelhuis et al.~\cite{middelhuis_2025} for the PPO agents and train each model with the same state and action space. For our proposed reward function (d), we use a discount factor of $\gamma=1$ since we aim to minimize the cumulative undiscounted return (compared to $\gamma=0.999$ for the other reward functions). Table~\ref{tab:reward_function_comparison} presents the optimality gap in the mean cycle time between the optimal policy and a PPO agent with different reward functions.

The code and trained models used to produce our results can be found in our repository: \url{https://github.com/jeroenmiddelhuis/BPO_rollouts}.

\begin{table}
\centering
\caption{Mean cycle time of the optimal policy and optimality gap (\%) between the optimal policy and PPO agent with different reward functions.}
\footnotesize{
\begin{tabular}{lcccccc}
\toprule
                 & \multirow{2}{*}{\makecell{Optimal \\ policy}}        & \multicolumn{4}{c}{Reward function}          & \multirow{2}{*}{Random} \\ \cmidrule{3-6}
                 &         & (a):  +1          & (b): $-CT_c$          & (c): $\frac{1}{1+CT_c}$     & (d): $-|C|(t-t^\prime)$  &  \\ 
\midrule
Low utilization  & 5.7 (0.05)   & 106\%      & 5927\%    & 2\%           & 4\%              &  18\%  \\
High utilization & 18.0 (0.55)  & 1384\%     & 4893\%    & 10\%          & 3801\%           &  101\% \\
Slow server      & 11.7 (0.23)  & 1940\%     & 7235\%    & \textbf{2\%}  & 6083\%           &  95\%  \\
Slow downstream  & 10.1 (0.20)  & 866\%      & 5629\%    & \textbf{0\%}  & 2422\%           &  15\%  \\
N-system         & 5.8 (0.06)   & 26\%       & 4822\%    & 34440\%       & \textbf{1\%}     &  12\%  \\
Parallel         & 9.7 (0.19)   & 1647\%     & 3419\%    & 15\%          & 41\%             &  16\%  \\ 
\bottomrule 
\end{tabular}
}
\label{tab:reward_function_comparison}
\end{table}

Table~\ref{tab:reward_function_comparison} shows that PPO's performance is highly sensitive to the reward function and scenario, as can be seen by the notable differences in performance. In contrast, our rollout-based algorithm and reward function, as shown in Table~\ref{tab:results_our_method}, can learn the optimal policy for all scenarios without the need for reward engineering, demonstrating their effectiveness and generalization capabilities.

Agents trained with reward functions (a) and (b) fail to outperform a random policy, which means that these reward signals are inadequate for effective learning. Reward function (a) does not account for the timing of case completions, and while the discount factor encourages earlier completions, it is not enough to learn a competitive policy. Reward function (b) considers timing but only provides penalties as a learning signal. Using this reward function, the agent may learn to avoid completing cases such that it receives mostly zero rewards.

On the other hand, an agent with reward function (c) outperforms the random policy and learns the optimal policy in two scenarios. In the N-system scenario, the agent hacks the reward function by completing only one of two activities to maximize rewards but not minimize the mean cycle time (see Section~\ref{sec:engineered_rewards}). A postpone penalty in \cite{middelhuis_2025} partially mitigates this, but the agent still did not learn a competitive policy. While this behavior is most evident in the N-system, similar undesired strategies can be learned in other processes.

Interestingly, with reward function (d), PPO learns the optimal policy in the N-system scenario but fails to find the optimal policy in the other scenarios. One reason is that our reward function (d) only provides negative rewards, making it difficult for the PPO agent to identify good actions, as all actions are penalized. Additionally, PPO works with episodic returns, so the cumulative reward depends on episode length. As a result, identical state-action pairs that occur later in an episode receive a lower cumulative reward than those that occur earlier, which makes it challenging to learn a good policy.

In conclusion, existing reward functions in combination with a PPO agent are not sufficient to learn competitive policies, while our method learns the optimal policy in all scenarios, outperforming the state-of-the-art in four scenarios and learning a similar policy in the other two scenarios.

\btext{
\subsection{Evaluation results on composite process models}\label{sec:composite_evaluation}
This section demonstrates our algorithm's capability to learn policies for increasingly complex process models, even as computational complexity grows.

The computational complexity of the algorithm can be measured in terms of the speed at which it learns a good policy, which depends on: the size of the neural network, consequently translating to size of state and action space, and the number of steps between taking an action and seeing the effect of that action (i.e. the moment at which a case completes). These factors are both determined by the number of tasks and the number of resources in a process model. Therefore, we gradually increase the size of a process model and observe the effect on the learning speed.

Specifically, starting with scenario 1, we add scenarios sequentially such that the departures of one scenario are the arrivals of the next, until we have a model that has the same complexity as the composite model from~\cite{middelhuis_2025}. We train the algorithm with the same hyperparameters only on the CTMDP, as it represents the original business process. We do not train an agent on the MDP as the results show no significant differences with different environment representations (Table~\ref{tab:results_our_method}). We removed the `postpone' action, because preliminary experiments revealed that it only added noise to the training dataset, and it was rarely selected as the best action. Finally, we report on the learning speed by measuring the number of learning iterations at which our model learns a policy competitive with the heuristics.

\begin{table}[]
    \centering
    \caption{\btext{Mean cycle time with 95\% confidence intervals for composite process models comparing our method against heuristic benchmarks, and the number of iterations required for our method to achieve performance equal to or better than the best heuristic.}}
    \btext{
    \footnotesize{
    \begin{tabular}{lcccccc} 
        \toprule
        Scenarios & Our method & \makecell{Iterations}  & SPT & FIFO & Random \\  \midrule
        1 & \textbf{5.6 (0.05)} & 1 & 5.9 (0.05) & 5.9 (0.06) & 6.7 (0.08) \\
        1-2 & \textbf{24.1 (0.67)} & 2 & 25.0 (0.53) & 32.7 (1.54) & 40.9 (1.73) \\
        1-2-3 & \textbf{36.7 (0.79)} & 1 & 52.7 (1.52) & 47.0 (1.70) & 60.9 (2.29) \\
        1-2-3-4 & \textbf{50.7 (1.00)} & 2 & 71.1 (1.67) & 55.8 (1.81) & 72.4 (2.46) \\
        1-2-3-4-5 & \textbf{57.6 (1.38)} & 3 & 81.7 (1.89) & 61.4 (1.72) & 76.9 (2.19) \\
        1-2-3-4-5-6 & \textbf{69.6 (1.19)} & 6 & 103.1 (2.56) & \textbf{70.8 (2.10)} & 87.6 (2.51) \\ 
        \bottomrule
    \end{tabular}}
    }
    \label{tab:results_composite_models}
\end{table}

Table~\ref{tab:results_composite_models} shows the results of the comparison. The table shows that for all levels of complexity, our method can learn a policy that performs at least as well as the best heuristic. While the computational complexity seems to increase exponentially, the number of learning iterations for these processes is still small and computationally manageable. In future work, further algorithmic improvements, such as those proposed by Temizöz et al.~\cite{TEMIZOZ2025}, and code optimization can still be applied to speed up the learning process.
}

%% file: 6_conclusion.tex
In this paper, we introduced a DRL method combining a rollout-based algorithm and a dense reward function to learn optimal policies for business processes. Our algorithm evaluates the execution trajectories of different actions in each state and uses the best-found state-action pairs to update the policy iteratively. Our reward function decomposes the sum of cycle times over the decision steps to provide a continuous feedback signal.

The results show that our proposed method consistently learns the optimal policy for the six evaluated business processes. In contrast, the current state-of-the-art algorithm learns the optimal policy in only two. \btext{In four of five composite processes, our method learns a policy that outperforms the best heuristic, and matches it in the fifth.} 
The benefit of using rollouts is that we can directly evaluate the outcome of different actions and do not need to approximate the value function, which is typical for other DRL methods~\cite{mnih2016asynchronous,schulman2017proximal}. Furthermore, a comparative analysis highlighted that while existing reward functions guide agents toward cycle time minimization, they can often be exploited, leading to suboptimal or unintended behaviors. On the other hand, maximizing our proposed reward function ensures that the objective is minimized, eliminating the need for problem-specific reward engineering.

In this paper, we presented a method to learn optimal policies for business processes. While we can compute the optimal policy on small business processes, we cannot do so for larger processes due to state explosion. \btext{In future work, the ability of our method to optimize real-world processes should be explored.} Furthermore, the business process with parallelism illustrated that the current state and action representation can be enhanced with case-specific features to learn better policies, which will also be investigated in future research.